\documentclass[conference]{IEEEtran}
\IEEEoverridecommandlockouts

\usepackage{amsmath,amsfonts,bm}

\def\eqref#1{equation~\ref{#1}}

\def\1{\bm{1}}

\DeclareMathAlphabet{\mathsfit}{\encodingdefault}{\sfdefault}{m}{sl}
\SetMathAlphabet{\mathsfit}{bold}{\encodingdefault}{\sfdefault}{bx}{n}

\def\gX{{\mathcal{X}}}

\def\sR{{\mathbb{R}}}

\newcommand{\E}{\mathbb{E}}

\usepackage{subcaption}

\usepackage{soul}
\usepackage{cite}
\usepackage{amsmath,amssymb,amsfonts}
\usepackage{graphicx}
\usepackage{textcomp}
\usepackage{balance}
\usepackage{xcolor}

\usepackage{hyperref}
\usepackage{multirow}
\usepackage{booktabs}       

\usepackage{mathabx,algorithm}
\usepackage{algpseudocode}
\usepackage{float}

\def\BibTeX{{\rm B\kern-.05em{\sc i\kern-.025em b}\kern-.08em
    T\kern-.1667em\lower.7ex\hbox{E}\kern-.125emX}}

\newcommand{\rev}[1]{{#1}}

\newcommand{\ac}[1]{{\textbf{AC}}}
\newcommand{\rone}[1]{{\color{teal} R1}}
\newcommand{\rtwo}[1]{{\color{magenta} R2}}
\newcommand{\rthree}[1]{{\color{orange} R3}}
\newcommand{\rfour}[1]{{\color{red} R4}}

\newtheorem{definition}{Definition}

\algnewcommand\algorithmicforeach{\textbf{for each}}
\algdef{S}[FOR]{ForEach}[1]{\algorithmicforeach\ #1\ \algorithmicdo}

\newcommand{\para}[1]{\vspace{0.5em}\noindent\textbf{#1}}
\DeclareMathOperator{\Grasps}{\mathbf{\mathcal{G}}}

\newcommand{\method}{\textsc{GRaCE}}

\newcommand{\grasp}{\mathbf{g}}
\newcommand{\obs}{\mathbf{o}}
\newcommand{\btheta}{\boldsymbol{\theta}}

\begin{document}



\title{\LARGE \bf GRaCE: Balancing Multiple Criteria to Achieve\\Stable, Collision-Free, and Functional Grasps}

\author{\IEEEauthorblockN{Tasbolat Taunyazov\IEEEauthorrefmark{1}, Kelvin Lin\IEEEauthorrefmark{1}, and Harold Soh\IEEEauthorrefmark{1},\IEEEauthorrefmark{2}}
\IEEEauthorblockA{\IEEEauthorrefmark{1}\textit{Dept. of Computer Science, National University of Singapore}}
\IEEEauthorblockA{\IEEEauthorrefmark{2}\textit{Smart Systems Institute, National University of Singapore}\\
Email: \{tasbolat, klin-zw, harold\}@comp.nus.edu.sg
}
}


\newcommand{\hsnote}[1]{[{\color{green}HS: {#1}}]}
\newcommand{\ttnote}[1]{{\color{magenta}TT: {#1}}}

\maketitle

\begin{abstract}
This paper addresses the multi-faceted problem of robot grasping, where multiple criteria may conflict and differ in importance. We introduce \rev{a probabilistic framework,} Grasp Ranking and Criteria Evaluation (\method{}), which employs hierarchical rule-based logic and a rank-preserving utility function \rev{for} grasps based on various criteria such as stability, kinematic constraints, and goal-oriented functionalities. \rev{\method{}'s probabilistic nature means the framework handles uncertainty in a principled manner, i.e., the method is able to leverage the probability that a given criteria is satisfied.}
Additionally, we propose \method{}-OPT, a hybrid optimization strategy that combines gradient-based and gradient-free methods to effectively navigate the complex, non-convex utility function. Experimental results in both simulated and real-world scenarios show that \method{} requires fewer samples to achieve comparable or superior performance relative to existing methods. The modular architecture of \method{} allows for easy customization and adaptation to specific application needs.
\end{abstract}

\section{Introduction}

\begin{figure}
\centering
	\includegraphics[width=.99\columnwidth]{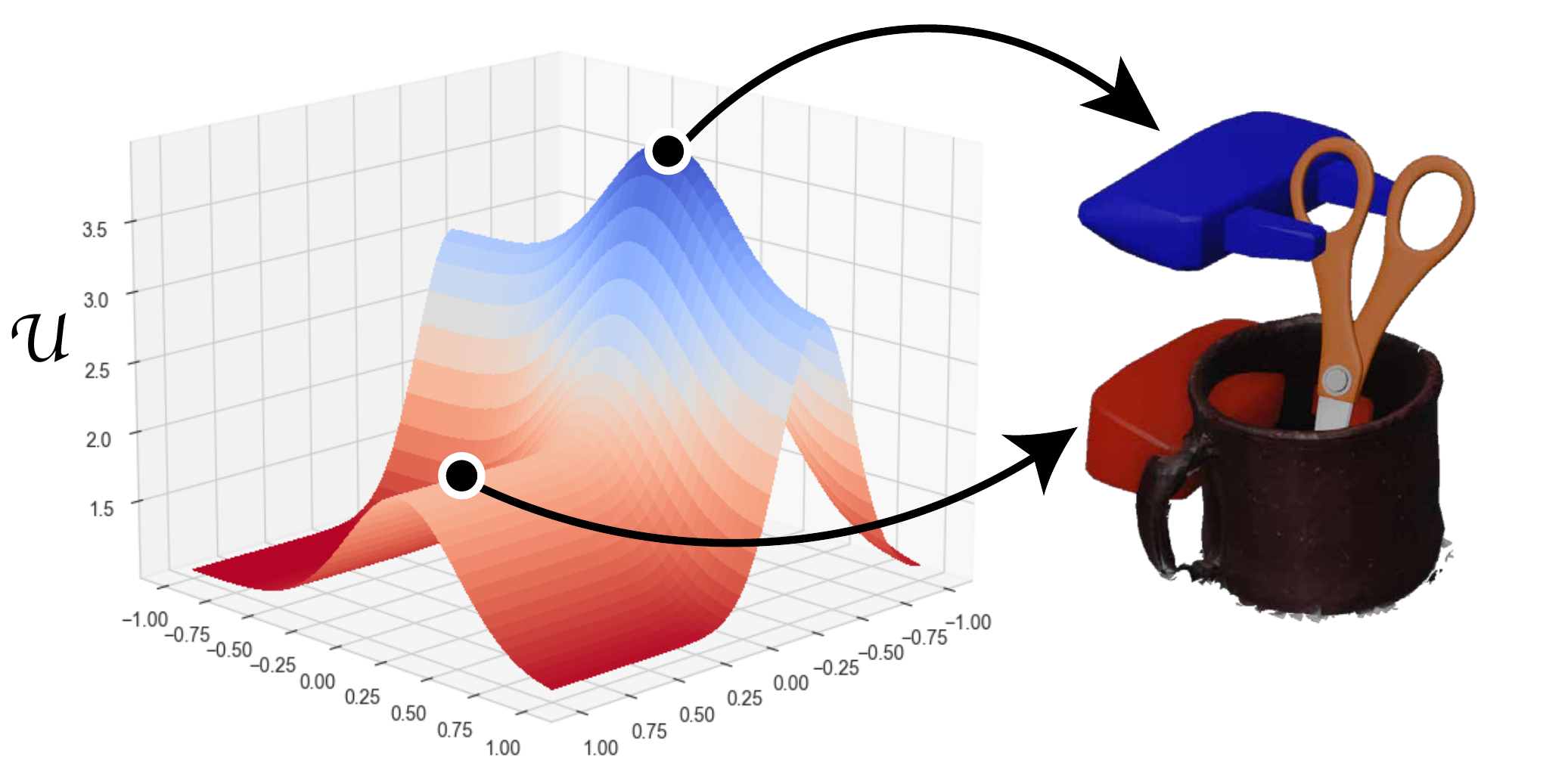}
	\caption{In this work, we formalize optimization of grasps under multiple ranked criteria. \rev{Our probabilistic framework, \method{},} defines an expected grasp utility $U$ where blue regions indicates higher utility values that are collision free and stable. We present a hybrid optimization method (\method-OPT) for finding grasps that maximize $U$.}
	\label{fig:main_fig}
\end{figure}

Grasping an object is typically influenced by the intended goal, which directly impacts the choice of grasp. For instance, we naturally grasp scissors by the handle for cutting, but by the blade when passing them safely to someone else. However, the goal isn't the sole or primary factor in determining a grasp; if the blade is obstructed or inaccessible, as illustrated in Fig. \ref{fig:main_fig}, we would opt to grasp the scissors by the handle, even if the intention was to hand it over. This scenario highlights that (i) multiple criteria influence the selection of a grasp and (ii) there exists a hierarchy of priorities among these criteria. The necessity for the grasp to be stable, accessible, and to avoid collisions with surrounding objects (like a mug) takes precedence over the functional goal of the grasp.

In this work, we are motivated by the problem of generating grasps that satisfy multiple criteria of differing importance. We apply hierarchical rule-based logic to robot grasping \cite{wilson1993hierarchical} and introduce a grasp utility function that is \emph{rank-preserving}~\cite{tuumova2013minimum}, i.e., it assigns larger utility values to grasps that satisfy higher ranked constraints. For example, robots are bound by their kinematic and dynamic constraints, which limits whether a proposed grasp can be performed, and environmental constraints (e.g., grasps should not collide with other objects). A stable grasp that satisfies these constraints should have larger utility than one that sacrifices these criteria for a functionally appropriate (but non-executable) grasp. 

We take a \emph{probabilistic} approach and optimize the \emph{expected utility} of a grasp, where the probability of a grasp satisfying a specific criteria is given by a classifier. Additional classifiers \rev{(whether data-driven or hand-crafted)} can be added (or removed) depending on the precise requirements of the application. 
This modular approach --- which we call \textbf{G}asp \textbf{Ra}nking and \textbf{C}riteria \textbf{E}valuation (\method{}) --- enables a robot to trade-off multiple conflicting criteria in complex contexts where not all desired objectives can be satisfied. \rev{\method{}'s probabilistic nature incorporates \emph{uncertainty} in a principled manner. This is crucial in real-world robotic scenarios, which often involve partial observability and noise. By utilizing the likelihood of a grasp meeting various criteria, \method{} moves beyond binary satisfaction assessments and reduces the risk of misclassification inherent in the true/false assignments used in prior work (e.g., \cite{veer2022receding}).}

\label{sec:intro}

\begin{figure}
\centering
	\includegraphics[width=.99\columnwidth]{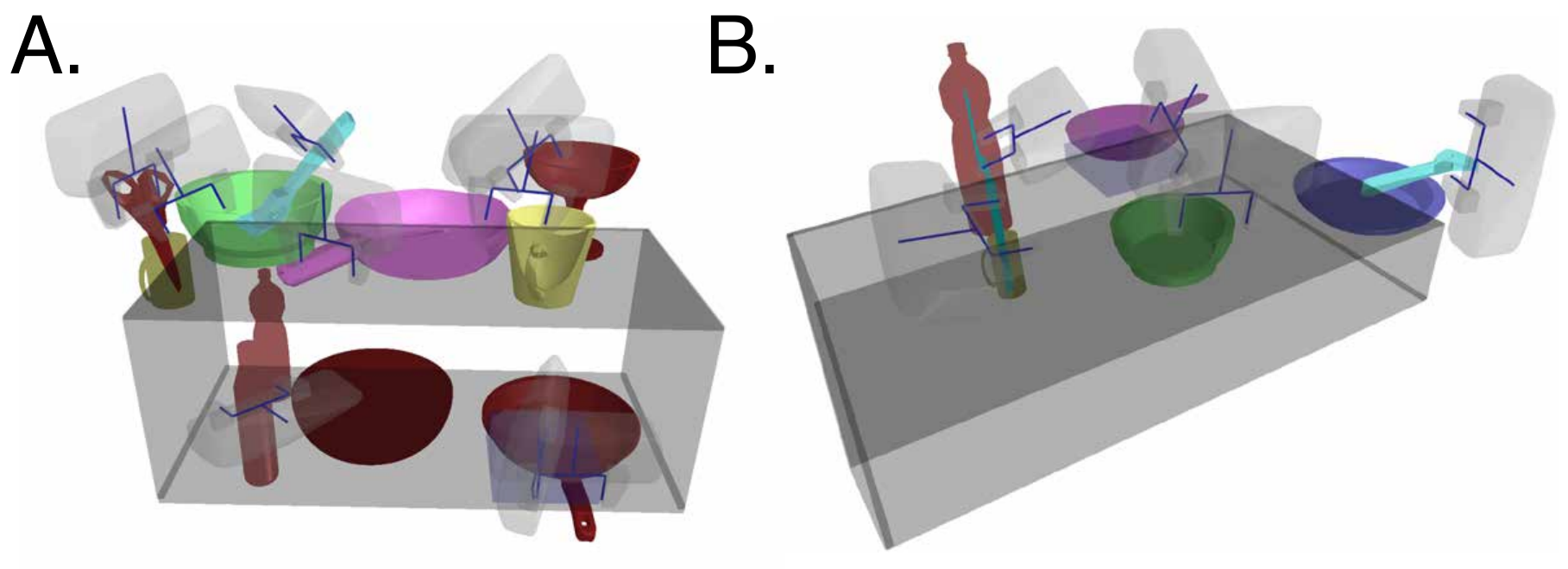}
	\caption{Shelf and Diner benchmark environments with sample grasps (in blue) of high utility.}
	\label{fig:environments}
\end{figure}

Although the utility function enables scoring of grasps, it is a complicated non-convex function to optimize, especially when the classifiers are themselves complex (e.g., a deep neural network). Inspired by progress in gradient-free methods (e.g., \cite{pourchot2018cem}), we propose \method{}-OPT --- a \emph{hybrid} optimization method that combines gradient-based and gradient-free optimization. Specifically, \method-OPT applies gradient-free optimization, an Evolutionary Strategy (ES)~\cite{beyer2002evolution}, to conduct a more ``diverse'' exploration over the landscape and prevent the optimization process from getting stuck at local optima. However, on its own, this gradient-free method can be slow to converge. As such, we use gradient-based optimization on a surrogate function, the  \emph{lower-bound} of the utility, to improve convergence speed. Experiments in complex environments show that \method{}-OPT requires significantly fewer samples and less time to achieve comparable (or better) performance to a filtering method used in prior works~\cite{ten2017grasp, liang2019pointnetgpd}. Our evaluations involved two simulated grasping scenarios --- shelf and diner (Fig. \ref{fig:environments}) --- in IsaacGym~\cite{makoviychuk2021isaac} and two real-world scenarios; these test scenarios are designed to be challenging (cluttered with small optimal grasping regions) and where the probability of satisfying multiple criteria may be traded-off. 

To summarize, this paper contributes \method{} which comprises a utility function that assigns higher values following user-specified hierarchical rules and an optimization method that uses both gradient-free and gradient-based optimization of the expected utility. Code and implementation details can be found online at \url{https://github.com/clear-nus/GRaCE}.

\section{Background and Related Work}
\label{sec:background}

\method{} is a \rev{probabilistic} framework for optimizing 6-DoF grasps. It builds upon related work on 6-DoF grasp candidate generation and prior work on the optimization of multiple criteria specified via rule/constraint hierarchies. We briefly review these topics in this section.

\para{6-DoF Grasp Filtering and Refinement.} Generating appropriate 6-DoF grasping remains an active area of research. One common approach is to first \emph{sample} a set of grasp, either through data-driven methods~\cite{mousavian20196,taunyazov2023refining}, heuristics~\cite{ten2017grasp} or a combination of both~\cite{liang2019pointnetgpd}, then \emph{filter} the grasps using evaluators to select the most promising candidates for execution. This sample-and-filter approach is common and can be very effective in practice~\cite{ten2017grasp}. However, it can be time-consuming in complex environments even with state-of-the-art samplers, especially the optimal grasp regions are small. 

An alternative approach to optimize grasps directly. Early work on multi-fingered end-effector grasping~\cite{zhou20176dof} demonstrated that a scoring function for grasp quality (a pre-trained classifier)  can be used to optimize grasps toward high quality regions. More recent work have applied optimization together with sample-and-filter methods, e.g., GraspNet~\cite{mousavian20196} optimizes/refines grasp samples using the quality evaluator. These methods focus on a single quality criterion, where else our work addresses the problem of trading-off multiple conflicting criteria. 

\method{} can also be seen as a contrasting approach to ``end-to-end'' data-driven 6-DoF grasping~\cite{sundermeyer2021contact, fang2020graspnet} where the sampler is trained to generate grasps that satisfy multiple criteria. However, these methods require retraining when a new criterion is added/removed, which is computationally expensive. \method{} enables the inclusion and removal of grasp criterion ``on-the-fly'', which we believe is more practical for many real-world applications. This aspect is similar to very recent work~\cite{taunyazov2023refining} that refines grasps using gradient flows, but \method{} enables the ranking of multiple criteria and we propose a hybrid optimization technique.




\para{Task/Functional Grasping.} 
Ensuring the functionality of a grasp is one of the most challenging aspects of grasp synthesis. Functional grasps depend on the semantic knowledge of the object and the target task. For instance, in a handover task involving scissors, the robot should grasp the blade so the human can take the handle. These grasps are referred to in the literature as functional~\cite{zhang2023functionalgrasp}, task-oriented~\cite{murali2020taskgrasp}, or semantic grasps~\cite{dang2014semantic, li2024semgrasp}.

Current methods for achieving task functional grasping are largely based on a sample-and-filter methodology with two-stage filtering. To elaborate, grasps are first sampled and non-stable ones are filtered out. Then, the remaining grasps are filtered based on appropriate affordances by segmenting objects~\cite{nguyen2016detecting,li2020learning,mandikal2021learning}. Deep learning methods are often used provide a grasp score given a grasp and a specific target task~\cite{murali2020taskgrasp}. Very recent methods employ Multimodal Large Language Models (LLMs) to reason about functional grasps~\cite{li2024semgrasp, chang2024text2grasp, mirjalili2023lan, tang2023task}. However, robustness of these models remains a challenge and research on LLMs for task-oriented grasps is ongoing.

The modularity of the GRaCE framework allows the use of many existing differentiable methods for assessing grasp functionality. In our work, we use  TaskGrasp~\cite{murali2020taskgrasp} as a criterion evaluator due to code availability and its good performance on various objects.

\para{Hierarchical Optimization of Multiple Criteria.}
A key component of our framework is a utility function, which leverages a rule hierarchy. Rule hierarchies have a long history in optimization, with early works dating back to 1967~\cite{waltz1967engineering}. More recent methods encode rule hierarchies using temporal logic~\cite{dimitrova2018maximum, tuumova2013minimum}. Unlike these methods, our framework is differentiable and we do not have to rely on external SAT solvers for optimization. Our work is closely related to very recent research on planning with a rank-preserving reward function for autonomous vehicles~\cite{veer2022receding}. Our grasp utility function has a similar structure to their proposed reward function, but our approach is probabilistic \rev{to handle uncertainty and} we optimize the expected rank of the grasp via a hybrid optimization method. 


\section{Ranking Grasps via Utility Functions}
In this section, we present our approach for trading-off criteria for grasp generation. A grasp $\grasp$ is typically defined as a set of contact points with an object which restricts movement when external forces are applied~\cite{bicchi2000robotic}. For simplicity, we will refer to end-effector poses as grasps (or grasp candidates) and denote them as $\grasp$, even if they do not satisfy the definition above (e.g., the pose does not make contact with the object). We first discuss how grasp criteria can be ranked, followed by our a utility function, and, finally, formulate an optimization based grasp generation method. 

\begin{table}
\centering
\caption{Formulas and Grasp Criteria with Associated Probability.}
\label{tab:graspopt_importance}
\small
\begin{tabular}{cll}
\hline
\rule[1ex]{0pt}{2ex} \textbf{Priority} & \textbf{Rule} & \textbf{Probability} \rule[-1ex]{0pt}{2ex} \\ \hline
\rule{0pt}{3ex} 1          & $\phi^{(1)} = \bigwedge_{j=1}^{M_1} c_{j}^{(1)}$   &  $\prod_{j=1}^{M_1} p_{j}^{(1)} $ \\
$\vdots$  \rule[-3px]{0pt}{3ex}  & \multicolumn{1}{c}{$\vdots$}            & \multicolumn{1}{c}{$\vdots$}     \\
$N$          & $\phi^{(N)} = \bigwedge_{j=1}^{M_N} c_{j}^{(N)}$  & $\prod_{j=1}^{M_N} p_{j}^{(N)}$ \rule[-2ex]{0pt}{2ex}  \\ \hline
\end{tabular}
\end{table}

\para{Criteria, Priority, and Rules.} We define a grasp criterion as a predicate $c^{(i)}_j(\grasp)$ where $i \in \{1,...,N\}$ is the criterion's priority (with descending importance) for a grasp $\grasp$ and $j$ is an index of criterion, $j = 1, \dots, M_i$. $M_i$ is a number of criteria with the same priority $i$. A rule $\phi^{(i)}(\grasp)$ is defined as a conjunction of criteria $\phi_i(\grasp) = \bigwedge_j^{M_i} c_{j=1}^{(i)}(\grasp)$. 

Let $p_j^{(i)} := P(c_j^{(i)}(\grasp)|\obs)$ be the probability that criterion $c^{(i)}_j(\grasp)$ is satisfied under observed context $\obs$. For notational simplicity, we will drop the explicit dependence of $p_j^{(i)}$, $c_j^{(i)}$ and $\phi^{(i)}$ on $\grasp$ and $\obs$. We assume that criteria are conditionally independent given the grasp and context. As such, the probability that a rule $\phi^{(i)}$ is satisfied is given by $\prod_{j=1}^{M_i} p_{j}^{(i)}$. Table~\ref{tab:graspopt_importance} shows a list of priorities, rules, and their associated probabilities.

\para{Rule Hierarchy and Rank of a Grasp.} A rule hierarchy $\psi$ is defined as a sequence of rules $\psi := \{\phi^{(i)}\}_{i=1}^N$. The rule hierarchy induces a total order on the grasps, enabling us to rank grasps. A grasp that satisfies all the rules has the highest rank, i.e., rank 1. A grasp that satisfies all the rules except the lowest priority rule has rank 2. This continues on, with grasps satisfying none of the rules having the lowest rank. Formally, we define a rank of a grasp as:

\begin{definition} Let $\psi$ to be rule hierarchy with $N$ rules. Let $\emph{eval} : \phi^{(i)} \mapsto \{0,1\}$ be a function that evaluates rule $\phi^{(i)}$ to be 1 if the rule is satisfied, 0 otherwise. Then the rank of the grasp $r : \Grasps \mapsto \mathbb{R}$ is defined as:
$$r(\grasp) := 2^N - \sum_{i=1}^N 2^{N-i} \emph{eval}(\phi^{(i)})$$
\end{definition}


\begin{table}
\centering
\caption{Rank-Preserving Grasp Utility}
\label{tab:graspopt_utility}
\small
\begin{tabular}{cll}
\hline
\rule[1ex]{0ex}{1ex} $r(\grasp)$  & \rule[1ex]{-3ex}{1ex} \textbf{Satisfied Rules} & \rule[1ex]{0ex}{1ex}  \textbf{Probability} \rule[-1ex]{0pt}{2ex} \\ \hline
\rule[1ex]{-2ex}{2ex} $1$ & $\bigwedge_{i=1} \phi^{(i)}$   & $\prod_{j=1}^{M_1} p_{j}^{(1)} \cdots \prod_{j=1}^{M_N} p_{j}^{(N)}$     \\
\rule[1ex]{0ex}{1ex}  $\vdots$ & $\vdots$            & $\vdots$      \\ 
\rule[1ex]{0ex}{1ex} $2^N$ & $\bigwedge_{i=1} \neg \phi^{(i)}$  & $(1-\prod_{j=1}^{M_1} p_{j}^{(1)}) \cdots (1-\prod_{j=1}^{M_N} p_{j}^{(N)})$ \rule[-2ex]{0pt}{2ex} \\ \hline
\end{tabular}
\end{table}


Table~\ref{tab:graspopt_utility} summarizes grasp ranks for the rule hierarchy and our utility is defined as the negative expected rank,   
\begin{equation}\label{eq:grasp_utility_total}
    U(\grasp) = -\E_{\psi} [ r(\grasp) ] = \sum_{i=1}^N 2^{N-i} \prod_{j=1}^{M_i} p_{j}^{(i)} - 2^N
\end{equation}
This simplified form can be obtained by observing that $\text{eval}(\phi^{(i)})$ is a Bernoulli variable with probability $\prod_{j=1}^{M_i} p_{j}^{(i)}$,
\begin{small}
\begin{align*}
    U(\grasp) & = -\E_{\psi} [ r(\grasp) ] & \\
    & = -\E_{\psi} [ 2^N - \sum_{i=1}^N 2^{N-i} \text{eval}(\phi^{(i)}) ] & \text{(by definition)} \\
    & = -2^N +\sum_{i=1}^N 2^{N-i} \E_{\psi}[\text{eval}(\phi^{(i)})] & \text{ (by linearity of } \E \text{)} \\
    & = -2^N + \sum_{i=1}^N 2^{N-i} \prod_{j=1}^{M_i} p_{j}^{(i)} & \text{}
\end{align*}
\end{small}

\para{Problem Statement.} We seek to find a grasp that maximizes the utility function:
\begin{equation}\label{eq:utility_maximization}
    \grasp^* = \arg\max_{\grasp} U(\grasp) = \arg\max_{\grasp} \sum_{i=1}^N 2^{N-i} \prod_{j=1}^{M_i} p_{j}^{(i)}
\end{equation}
The key challenge is that Eq.~(\ref{eq:utility_maximization}) is a non-convex function of the grasps that can trap standard gradient-based methods. Moreover, the multiplication of probabilities leads to numerical instabilities with vanishing gradients when used with deep neural classifiers~\cite{hochreiter1998vanishing, hanin2018neural}. In the next section, we describe how to optimize this function using \rev{\method{}-OPT}.



\section{Hybrid Optimization of Grasps}

In this section, we introduce \method-OPT, a hybrid optimization technique that leverages both gradient-free and gradient-based methods to optimize Equation~(\ref{eq:utility_maximization}). As an initial step, we considered optimizing a lower-bound of Eq.~\ref{eq:utility_maximization} using Jensen's inequality:
\begin{align*}
    \log U(\grasp) &= \log \left( \frac{N}{N} \sum_{i=1}^N 2^{N-i} \prod_{j=1}^{M_i} p_{j}^{(i)}\right) \nonumber  \\
    & = \log N + \log{ \left(\frac{1}{N} \sum_{i=1}^N 2^{N-i} \prod_{j=1}^{M_i} p_{j}^{(i)}\right)} \\
    &\geq \log N + \frac{1}{N} \sum_{i=1}^N  \log \left( 2^{N-i} \prod_{j=1}^{M_i} p_{j}^{(i)} \right)  \nonumber \\
    &=\log N +\frac{1}{N} \sum_{i=1}^N (N-i) \log 2 +\frac{1}{N}\sum_{i=1}^N \sum_{j=1}^{M_i} \log p_{j}^{(i)} \\
    & = C + \frac{1}{N}\sum_{i=1}^N \sum_{j=1}^{M_i} \log p_{j}^{(i)} \triangleq L(\grasp)
\end{align*}
where $C$ is a constant independent of $\grasp$. Empirically, we find $L(\grasp)$ to be easier to optimize and numerically stable, but inspection of its form shows that it is no longer rank preserving since the utilities are factored out. 

As such, we only use this gradient-based optimization as an inner-loop within a gradient-free ES setup~\cite{beyer2002evolution}, shown in Algorithm \ref{alg:graspoptes} below.  We assume that we have access to a grasp sampler $q_0$ from which we can sample initial grasps $\mathbf{G}_0$ from (line 1). In practice, $q_0$ can be any grasp candidate sampler, e.g., GraspNet-VAE~\cite{mousavian20196} or a heuristic sampler such as GPD~\cite{ten2017grasp}. We then optimize these grasps over $T$ outer gradient-free iterations (lines 2-10). In detail: new batches of grasps are sampled using a multivariate Gaussian distribution with mean $\grasp_t$ and covariance matrix $\Sigma$. The covariance $\Sigma$ is manually selected in our experiments but can also be adaptive~\cite{igel2007covariance}. Lines 4 to 6 optimizes the lower bound $L(\grasp)$. Line 8 and 9 assesses grasps using $U(\grasp)$ and selects the top $R$ grasps. In preliminary experiments and ablations (Sec. \ref{sec:ablations}), we found \method-OPT to be superior to using either gradient-based or gradient-free methods alone. 


\begin{algorithm}
\caption{\method-OPT}
\label{alg:graspoptes}
\begin{algorithmic}[1]
\Require grasp sampler $q_0$, utility $U(\grasp)$, lower bound for utility $L(\grasp)$, number of update steps ($T$), covariance matrix ($\Sigma$), size of the local grasps ($\Theta$), step size ($\eta$), number of update steps for lower bound ($K$), size of the grasps ($R$). 
\State $\mathbf{G}_1 \gets \{ \}$
\For{$r \gets 1$ to $R$}  
    \State $\grasp_r \sim q_0$ \Comment{Sample initial grasp candidates}
    \State $\mathbf{G}_1 \gets \mathbf{G}_1 \cup \{ \grasp_r \}$
\EndFor
\For{$t \gets 1$ to $T$}
    \ForEach{$\grasp_r \in G_t$}
    \For{$\theta \gets 1$ to $\Theta$}
    
    \State $\grasp_r^{(0)} \sim \mathcal{N}(\grasp_r ,\Sigma)$ 
    \For{$k \gets 1$ to $K$} \Comment{Optimize new samples}
        \State $\grasp_r^{(k)} = \grasp_r^{(k-1)} + \eta \nabla_{\grasp_r^{(k-1)}} L({\grasp}_r^{(k-1)})$
    \EndFor
    \State $\mathbf{G}_t \gets \mathbf{G}_t \cup \{ \grasp^{(K)}_r \}$
    \EndFor
    \EndFor


    \State
    $\mathbf{G}_t \gets \text{topR}(\mathbf{G}_t)$ \Comment{Select $R$ grasps with highest $U$}

\EndFor
\State \Return $\mathbf{G}_T$ \Comment{Optimized grasps}
\end{algorithmic}
\end{algorithm}


\section{Criteria for successful 6-DoF grasps}
In this section, we describe different grasp criteria used in our experiments. We assume a setup where a human user is asking the robot to perform a task, e.g., to ``handover the scissors''. The robot has access to the natural language utterance from the human as well as observations of the environment (a point cloud). 

As previously mentioned, we assume a probabilistic setup where the probability of criteria satisfaction is given by a classifier $P(c^{(i)}_j(\grasp) | \obs)$. We used four different classifiers that capture different quality aspects of a grasp: stability, executability, collision-free, and functional. We discuss these classifiers at a high-level and leave implementation details to the \rev{online code repository (made public if accepted)}.

\para{Stability Classifier (S).} We use the  stability evaluator in~\cite{taunyazov2023refining}. The classifier takes as inputs a grasp pose and a point cloud of the object, and outputs a prediction of grasp stability.

\para{Execution Classifier (E).}
Our execution classifier captures two important aspects of robot poses: reachability map~\cite{makhal2018reuleaux} and kinematic singularity~\cite{rubagotti2019semi}.
\begin{figure}
\centering
\includegraphics[width=0.4\columnwidth]{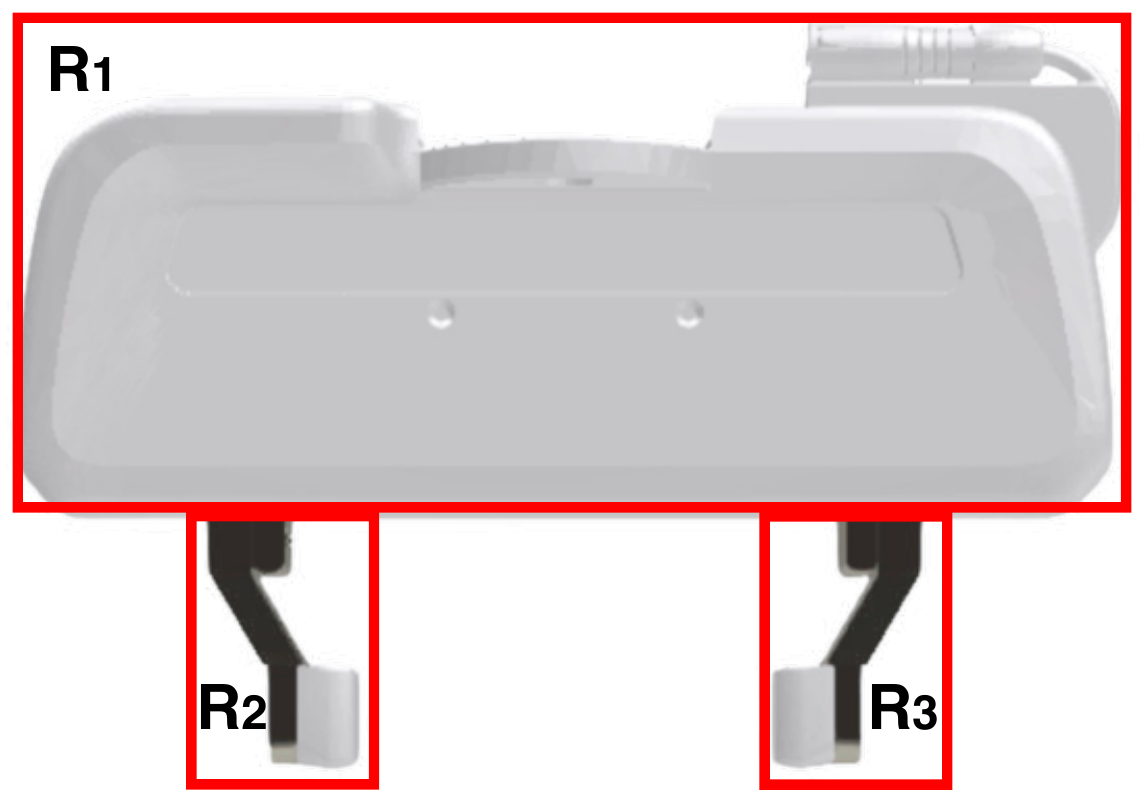} 
    \caption{Convex decomposition of the gripper used for Collision Detection Classifier.}
    \label{fig:gripper_convex}
\end{figure}
We calculate the manipulability score for a given grasp:
\begin{equation} \label{eq:vol_man_ellipsoid}
    \omega(\btheta) = \sqrt{\det{\mathbf{J(\btheta)J(\btheta)^T}}} \geq 0
\end{equation}
where $\mathbf{J(\btheta)}$ is the Jacobian matrix and $\btheta$ is the Inverse Kinematics (IK) solution. Then, we define the predicted grasp pose, $\Tilde{\grasp}$, using the IK solution:
$$\Tilde{\grasp} = \text{FK}(\btheta)$$
Finally, we combine this two quantities to yield,
\begin{equation}\label{eq:execution_classifier}
    p(\text{eval}( c_e(\grasp) )=1) = 
    \begin{cases}
    \sigma( -C_m d(\grasp, \Tilde{\grasp} )),& \text{if } \rev{d(\grasp, \Tilde{\grasp}) < d_{\epsilon}} \\
    \sigma(C_w(\omega({\btheta}) - \omega_{\text{th}})),              & \text{otherwise}
    \end{cases}
\end{equation}
where \rev{$\sigma(z) = \frac{1}{1+ \exp^{-z}}$ is the logistic sigmoid}, $C_m$ and $C_w$ are scaling coefficients, $\omega_{\text{th}}$ is a lowest manipulability threshold that allows safe grasp execution, $d(\cdot,\cdot)$ is a distance function between predicted grasp pose from the IK solution and current pose calculated in SE(3)~\cite{belta2002euclidean} and \rev{$d_{\epsilon}$ is a IK tolerance}.

\para{Collision Detection Classifier (C).}
The backbone of our Collision Detection Classifier is the 3-D Signed Distance Function (SDF)~\cite{chan2005level}. For simplicity, we use the original version of SDF that is designed for convex objects. Let  $\gX \in \sR^{K \times 3}$ represent a point cloud represented with respect to the world frame with $K$ points and $\mathbf{x}_k \in \gX$ be a point within $\gX$. The SDF for the box $R_i$ is defined as 
\begin{equation}
    d_{R_i} = \frac{1}{|\gX|}\sum_{k=1}^{k=K} \| \max ( |\mathbf{x}|- \mathbf{H}, 0) \|_2 
\end{equation}
where $\mathbf{H} \in \sR^3$ is the half-extent of the box in Cartesian coordinates. We decompose the gripper into three boxes $R_1$, $R_2$ and $R_3$ as shown in Fig.~\ref{fig:gripper_convex}.The SDF is differentiable and we use it to create our collision detection classifier:
\begin{equation}
    P(\text{eval}( c_c(\grasp) )=1 | \obs) = \sigma\left(C_c (d_{\text{th}}-\frac{1}{3} \sum_{i=1}^{i=3} d_{R_i})\right)
\end{equation}
where $d_{\text{th}}$ is a user-defined threshold and $C_c$ is a scale coefficient.

\para{Intention Classifier (N).} Our intention classifier outputs the probability that the grasp location coincides with their intent.
We first extract the user's intent (e.g., ``Handover'') from their utterance (e.g., ``Hand over the knife'') using JointBERT~ \cite{jointbert}. Our JointBERT model is trained on a curated dataset of programmatically generated queries and evaluated on sentences surveyed from test users. To evaluate if the grasp matches the intention, we use TaskGrasp~\cite{murali2020taskgrasp} as it can identify affordance-rich and affordance-poor regions of objects. TaskGrasp evaluates grasps with respect to the point   cloud and task, and outputs  $P(\text{eval}( c_n(\grasp) )=1  | \obs)$. 
As TaskGrasp inherently assumes that all grasps are stable before inference, we lower the score to zero if the grasp is more than 3cm away from the nearest point in the point cloud; we find that this modification helps to reduce false positives.


\para{Summary and Ranking.} The above classifiers are all differentiable and gradients can be obtained using modern auto-differention libraries such as PyTorch \cite{imambi2021pytorch}. 
In our experiments, we rank the criteria as follows: the S-classifier has rank 1, the E-classifier and C-classifier have rank 2, and the N-classifier has rank 3. 
\begin{figure*}
\centering
        \includegraphics[width=0.95\textwidth]{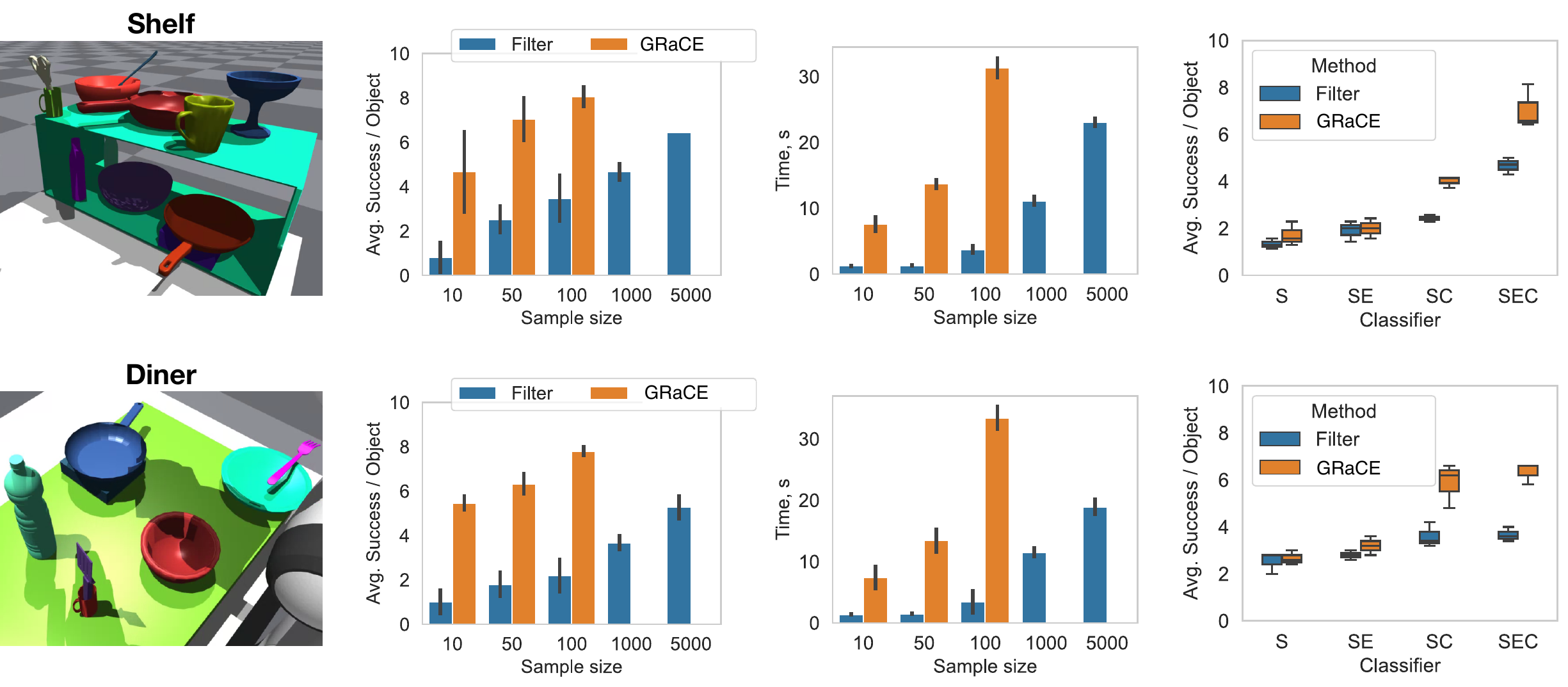}
    \caption{Results on Experiments on the Shelf (top) and Diner (bottom) Environments. The bar graphs show averages with standard deviation as error-bars. Using 50 samples, \method{} outperforms Filter (5000 samples) and takes less computational time.}
\label{fig:experiments}
\end{figure*}

\section{Simulation Experiment}
\label{sec:experiments}

The goal of our experiments is to establish if using the \method{} framework (i) yields suitable grasps in a complex environment, and (ii) trade-off multiple criteria. Moreover, \rev{\method{}-OPT} is more computationally expensive compared to a simple sample-and-filter approach (principally due to gradient computation). Is this added cost justified? Moreover, are the multiple criteria necessary for finding successful grasps and if so, can they be traded-off effectively? \rev{Our experiments are aimed at answering these questions.} \rev{To simplify exposition, we will refer to the process of using \method{}-OPT to optimize the expected utility in (\ref{eq:utility_maximization}) as \method{}.} 

\subsection{Simulated Environments}
\rev{We used IsaacGym from NVIDIA \cite{makoviychuk2021isaac}, which is a state-of-the-art simulator capable of simulating robot movement and object grasping.} We engineered two simulation environments, namely a Shelf module and a Diner module:
\begin{itemize}
    \item \textbf{Shelf} consists of a two-layered shelf with common everyday items placed on both layers. The Shelf module is designed to be cluttered and complex. Hence, the optimal grasping region for each object is confined to a small area, reducing the effectiveness of sampling-based methods.
    \item \textbf{Diner} consists of items that may be present in a typical dining setup, such as bowls, forks, a pan, and a spatula. 
\end{itemize}
The graspable objects in these environments are from the ShapeNet dataset, and the shelves and tables were created using Blender. We packaged the Shelf and Diner modules as a set of OBJ files that can be loaded into any simulator capable of importing OBJ meshes. 

\subsection{Experiment Process}

\para{Perception.} We first record point cloud data through IsaacGym's simulated depth and segmentation cameras from multiple views, and segment out the target object from the environment. 

\para{Grasp Sampling and Optimization.} GraspNet VAE~\cite{mousavian20196} is then used to sample grasps and optimized with \method{}. The resulting output is a list of grasp poses, along with their utility scores. 

\para{Hyperparameters.} The scaling coefficients for the E and C classifiers are set at $C_m=C_\omega=C_c=1000$ to approximate the behavior of a differentiable Heaviside function. A minimum manipulability threshold is established at $\omega_{t\text{th}}=0.001$, below which the robot's operational capability is compromised. The collision threshold is defined as $d_{\text{th}}=0.0025$, aligning with the planner's tolerance level. The learning rate for \method{}-OPT is determined through trial and error, set at $0.0001$ for translational adjustments and $0.00001$ for rotational adjustments.


\para{Grasp Planning, Execution, and Evaluation.} The optimized grasps are passed to Moveit!~\cite{chittamoveit} to generate trajectory plan. To minimize collision, instead of planning to the grasp pose, we plan the trajectory to a pre-grasp configuration 5cm linearly behind the actual grasp pose. The robot performs the grasping by moving the end-effector towards the object and closing its grippers. 
To execute the plan, we use a configuration-space controller to closely mimic and execute the planned trajectories. 
As IsaacGym is deterministic  across sessions, only one execution attempt of each trajectory was performed.
A grasp was termed as \emph{successful} if the target object is held by the gripper fingers after the trajectory was executed. Note that this measure of success excludes the intention criteria (which is subjective and handled separately). 

\para{Baseline Methods.} We compared the following methods: 
\begin{itemize}
    \item \method{} as described with all four classifers. 
    \item \method{} with only the S-classifier. This ablation uses a single criteria (stability) and is similar to the refinement used in GraspNet~\cite{mousavian20196} and its variants. This baseline serves to represent grasping methods where only a single criteria is applied. 
    \item Ablations of \method{} by removing  criteria, e.g., SE denotes that only the stability and execution classifier were used. These ablations enable us to see if excluding important criteria leads to more failures. 
    \item Sample-and-Filter, termed as ``Filter'', initially samples candidate grasps and subsequently filters them according to a predefined threshold of the grasp utility function. When used with the GraspNet VAE (as in our experiments), this baseline represents a modern deep-generative approach to grasping that allows for multiple criteria. 
\end{itemize}




\begin{figure}
\centering
\includegraphics[width=.86\columnwidth]{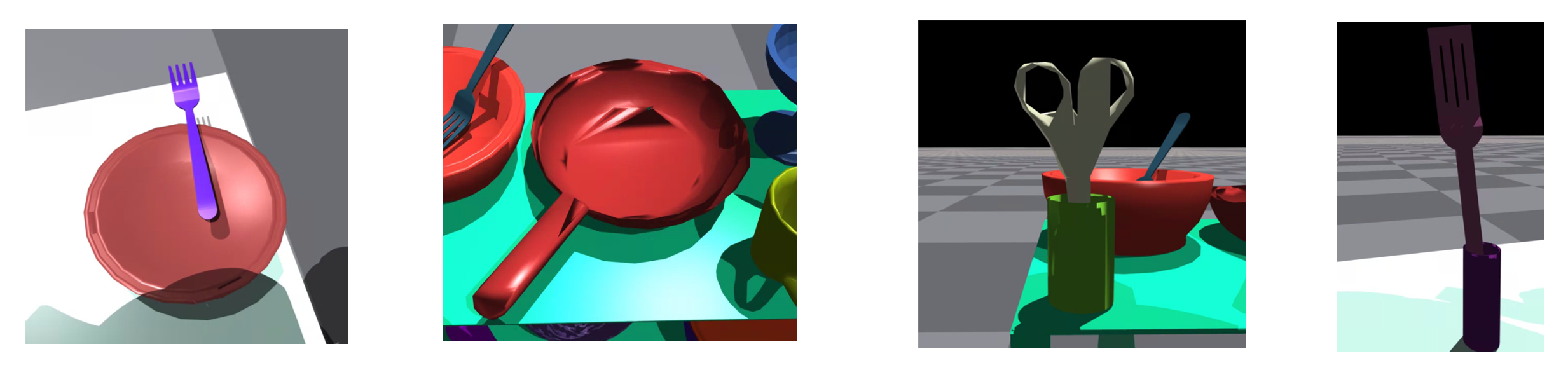}
	\caption{\rev{Selected objects for  intention evaluation: a fork, pan, scissors, and spatula}.}
	\label{fig:N_classifier_objs}
\end{figure}
\begin{figure}
\centering
\includegraphics[width=.86\columnwidth]{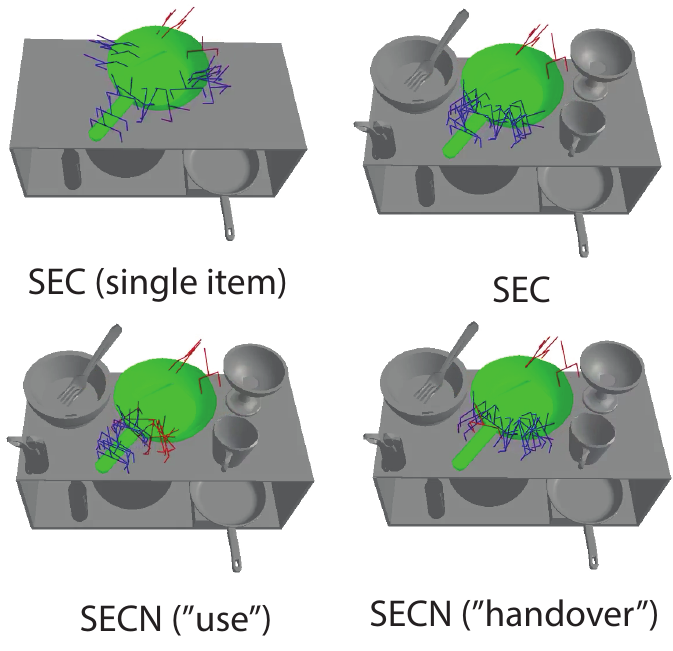}
	\caption{Case studies involving different grasp criteria for the pan on the shelf. Successful grasps are optimized based on the environment's collision-free space and the user's intent. \rev{Grasps are colored by their expected utility from red (low utility) to blue (high utility).}}
	\label{fig:goes_diff_types}
\end{figure}

\subsection{Results}
\label{ref:results}
In this section, we summarize our main findings. In general, we find \method{} to be superior to filtering on both the Shelf and Diner scenarios. Moreover, it is able to prioritize important criteria to find higher utility grasps. 

\para{Is optimization really necessary? Does \method{} outperform Sample-and-Filter? } 
We evaluated \method{} (SEC) against sample-and-filter with different sample sizes (10, 50, 100). Fig. \ref{fig:experiments} shows the average number of successes per object for the top-10 grasps across the different objects in the Shelf and Diner environments (seven and five objects, respectively). Note that the intention criterion was excluded as compliance with user intent involves subjective evaluation.

In Fig. \ref{fig:experiments}, we observe that \method{} outperforms Filter across the same sampling sizes (10, 50, 100). We further ran Filter with larger sample sizes (1000 and 5000), which enabled it to achieve attain similar performance to \method{}. At 5000 samples, Filter performs similarly to \method{} using 50 samples. However, this also resulted in it requiring almost 2x longer compute times. In short, although \method{} is more expensive \emph{per sample}, it is able to achieve better grasps with fewer samples. 

\para{Can \method{} optimize multiple criteria to find successful grasps?} The results of our \method{} ablations are shown in Fig. \ref{fig:experiments}. We observe that the using all three classifiers (SEC) resulted in the best performance. The marked increase in performance from SE to SC  may be attributed to the cluttered nature of Shelf and Diner, where many candidate grasp poses can collide with other objects. Qualitatively, the successful grasps for the pan (shown in blue  in Fig.~\ref{fig:goes_diff_types}) are biased towards collision-free areas when other objects are on the shelf objects.

\begin{figure}
    \centering
\includegraphics[width=0.86\columnwidth]{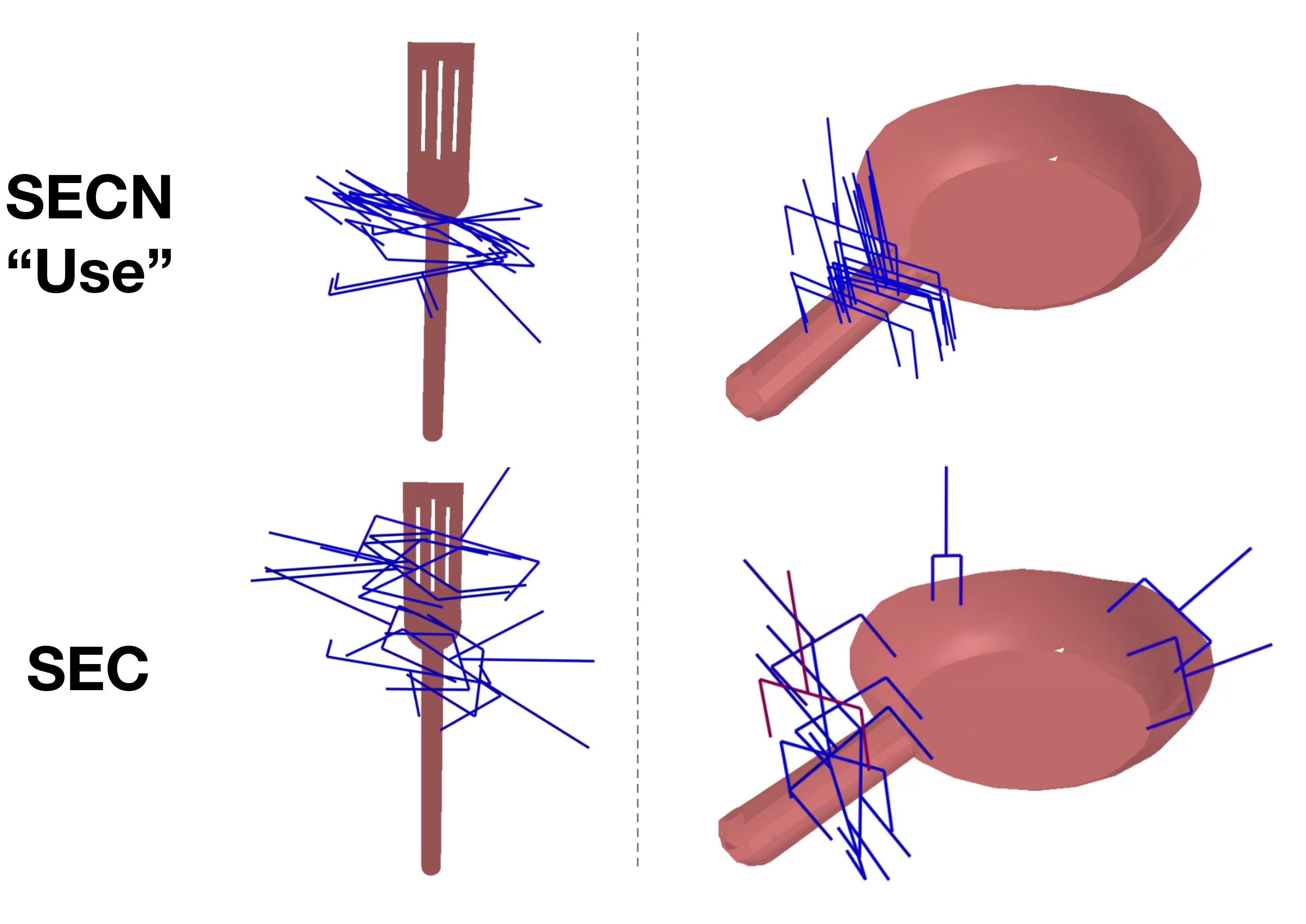}
    \caption{Incorporating the intention classifier (SECN) shifts grasps towards functional regions.}
    \label{fig:spatula_pan_grasp}
    \vspace{-0.5em}
\end{figure}
\begin{figure}
    \centering
\includegraphics[width=0.86\columnwidth]{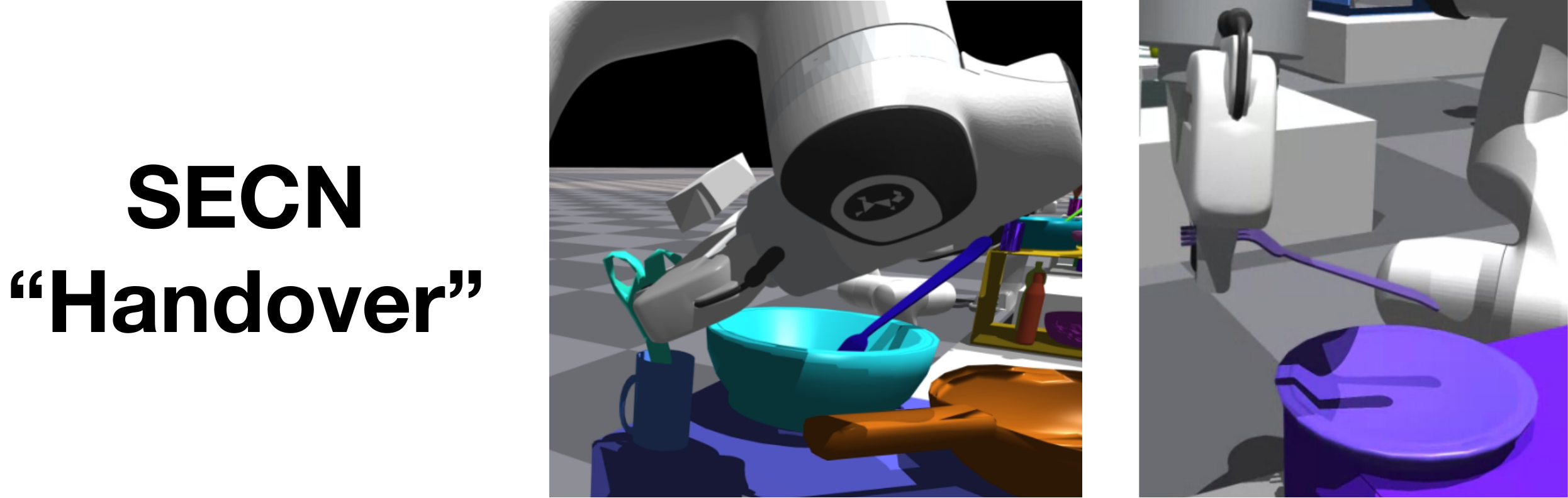}
    \caption{\method{} produces grasps that prioritize the higher ranked criteria, automatically sacrificing functional regions for stable, executable, and collision free grasps. \rev{In the fork example (right), \method{} generates a grasp that satisfies all criteria. However, in the scissors example (left), the correct part to grasp for handover is the blade but this would result in collision. Hence, \method{} picks a stable collision-free grasp instead.}}
    \label{fig:fork_scissors}
\end{figure}

\begin{figure*}[hbt!]
\centering
\includegraphics[width=0.95\textwidth]{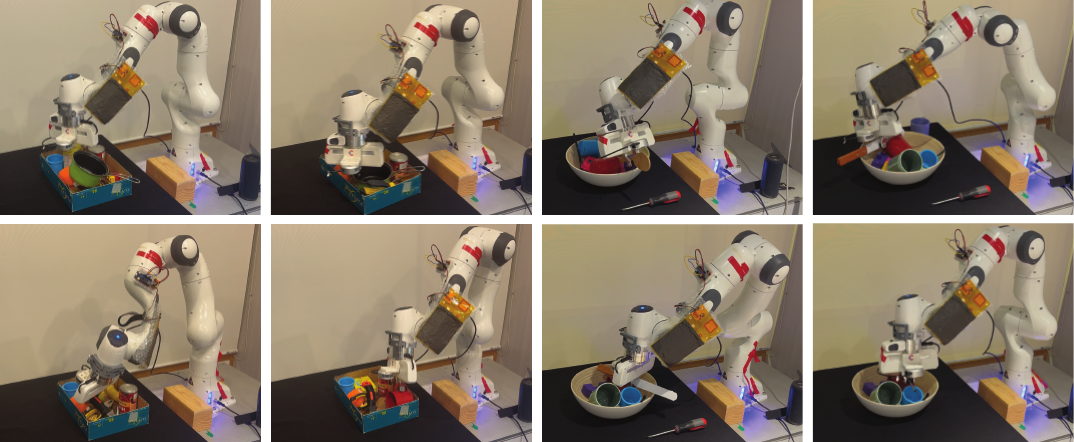}
	\caption{Successful 6-DoF grasps for real-world experiments with Panda robot for various items in the box and bowl scenarios.}
	\label{fig:goes_collage}
\end{figure*}

\para{Does \method{} with the intention classifier generate successful functional grasps?}
More precisely, we sought to evaluate if (i) \method{} would generate grasps in regions matching the user intent if the higher-ranked criteria can be satisfied, and (ii) prioritize the higher-ranked criteria, even if the resulting grasp has violates the the functional criteria. To that end, we selected four objects (shown in Fig. \ref{fig:N_classifier_objs}) and paired with pan and spatula with the ``Use'' intention, and the scissors and fork with the ``handover'' intention. 


Fig. \ref{fig:spatula_pan_grasp} shows the grasps generated with and without the intention criteria. To elaborate, the spatula can be separated into two regions: handle, which is ideal to grasp for ``use'', and the head, which should not be grasped for this purpose. Notably, both of these regions satisfy the stability, executable, and collision-free criteria. We see that \method{} using only SEC generated grasps in both regions, while \method{} with SECN produced grasps only at the handle. Similar grasps can be observed for the pan. Turning our attention to the ``handover'' intent, the scissors and fork are in placements that limit access to regions that have coincide with the ``handover'' intention. In this case, we observe \method{} (with SECN) to forgoes these regions and instead produces grasps that satisfy the other, more highly ranked, criteria (examples in Fig. \ref{fig:fork_scissors}). In Fig.~\ref{fig:goes_diff_types}, the ``use'' and ``handover'' intentions bias grasps toward the handle and body of the pan, respectively.

\section{Real-world Experiments}
\label{sec:realworld}

Thus far, we have discussed \method{} in simulation settings, but does \method's performance carry over to the real world? 
We conducted real-world tests comparing \method{} against the filter baseline, similar to the simulation setup. 

\para{Experimental Setup.} We use a Franka Emika Panda with a RGB-D camera (Intel RealSense L515~\cite{lourencco2021intel}) attached to the last link of the robot. The general execution pipeline for the real-world experiments are similar to the simulations except for perception. We use Detectron2~\cite{wu2019detectron2} to segment out the mask for our object of the interest. Then, we apply this mask to the corresponding depth map, and calculate the point clouds using the camera's intrinsic and extrinsic matrices. We also use built-in filter functions of Open3D library~\cite{zhou2018open3d} to remove outliers for pointcloud. During our experiments, we find that this method of processing the pointcloud resulted in more robust points compared to data driven methods such as Unseen Object Clustering~\cite{xiang2020learning} and Squeezesegv3~\cite{xu2020squeezesegv3}.

\para{Scenarios.} We tested our setup to grasp objects in two different scenarios:
\begin{itemize}
\item \textbf{Box}, where the robot was tasked to generate grasps for 10 different items in a clutter. Here, there is no intention criteria and the goal was to execute a stable grasp and lift the object. 
\item \textbf{Bowl}, where the robot attempted to grasp one of three different items (a wooden spoon, a knife, or a screwdriver) with the intent to handover the object. A grasp was successful if the robot managed to lift the object out of the bowl and hand it over to the experimenter.
\end{itemize}
Both these settings are challenging due to (i) noisy perception and (ii) the  feasible grasp region for each object was generally small due to the clutter.
In each experiment, we conducted 10 trials for each object to be grasped and recorded the number of grasp successes; in total, our experiment involved 260 real-world grasps. We set \method{} to use 50 samples, while Filter used 1000 samples. Both methods have comparable timings; \method{} took an average of $14$ seconds to obtain a grasp, while Filter took $12$ seconds. Note that these timings are for un-optimized Python implementations and future work can look into reducing this computation time.

\begin{figure}
\centering
\includegraphics[width=.95\columnwidth]{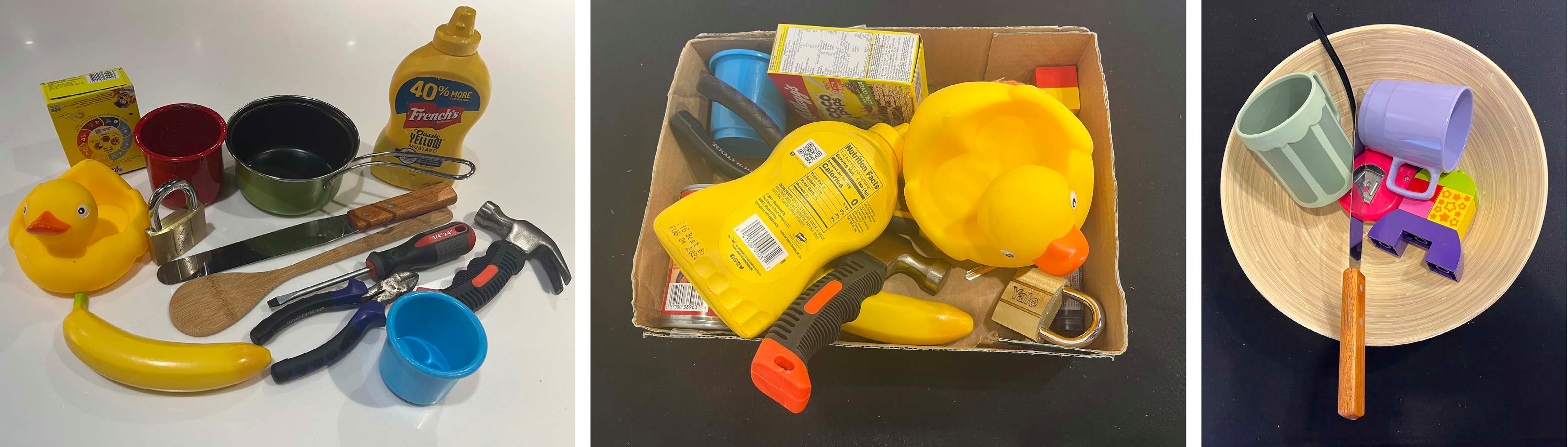}
	\caption{Real world items, along with the box and bowl scenarios.}
	\label{fig:real_world_exps}
\end{figure}

\begin{table}
    \centering
    \caption{Real-World Grasp Experiments: Average Success Rates with Standard Deviation in Brackets.}
    \normalsize
    \label{tab:realworld}
    \begin{tabular}{lcc}
    \toprule
         Method &  Box & Bowl  \\
        \midrule
         \method{} & 65\% (0.10) & 57\% (0.06)\\
         Filter & 31\% (0.12) & 33\% (0.06)\\
         \bottomrule
    \end{tabular}
\end{table}

\para{Does \method{} find successful multi-criteria grasps in real-world scenarios?} Our results, summarized in Table \ref{tab:realworld}, show that \method{} outperforms Filter in both domains by a significant margin. In both cases, \method{} achieves approximately double the success rate of Filter. Fig.~\ref{fig:goes_collage} depicts various successful 6-DoF grasps in the real-world experiment for the box and bowl scenarios. Qualitatively, we found \method{} to more reliably return a feasible grasp; in contrast, Filter failed to return \emph{any} suitable grasp in 26 out of the 130 trials (20\%). Other failures in both cases were commonly due to perception errors and robot trajectories executed near singular configurations, leading to grasp offsets, collisions, and robot errors (see Table \ref{table:failurecases}). Overall, our findings affirm that \method{} sustains its performance in real-world conditions. 

\begin{table}
\caption{Breakdown of Grasp Failure Types in Real-World Experiments. We recorded three kinds of failures: (i) Robot Errors (RE) where the robot fails to plan to the target grasp or stops due to a singularity, (ii) Grasp Failure (GF) refers to cases where the robot collides with an object or the environment, or grasps nothing, and (iii) No Grasp (NG) occurs when no sufficiently good grasp was generated (all expected utilities were below $0.01$)
}
\label{table:failurecases}
{\small
\begin{tabular}{l|ll|ll}
\hline
              & \multicolumn{2}{c|}{Box}                      & \multicolumn{2}{c}{Bowl}   \\ \hline
              & Filter                          & GRaCE       & Filter      & GRaCE        \\ \hline
RE   & \multicolumn{1}{r}{16\% (0.03)} & 31\% (0.03) & 12\% (0.06) & 20\% (0.014) \\
GF & 58\% (0.12)                     & 69\% (0.08) & 50\% (0.1)  & 30\% (0.07)  \\
NG      & 26\% (0.12)                     & 0\% (0.00)  & 38\% (0.06) & 50\% (0.07) \\ \hline
\end{tabular}
}
\end{table}

\rev{
\section{Further Ablation Experiments}
 \label{sec:ablations}
 
In this section, we report on additional ablation experiments designed to evaluate changes in priorities affect outcomes, and the effect of \method-OPT's  hyperparameters (specifically, the number of steps $T$ and lower-bound update steps $K$). In the following experiments, we re-ran our simulation experiment in the Shelf environment with the Stability (S), Execution (E), and Collision (C) criteria. 

\para{Does changing criteria priorities alter the resultant grasps?} In this ablation, we changed the priorities of the Execution and the Collision criteria. Table \ref{table:changecriteria} shows the average scores of the grasps for each of the different criteria. Compared to the initial sampled grasps, \method{} improves scores differently depending on the specified priorities. For example, when C was prioritized over E (S$>$C$>$E), the average score for the collision criteria was 0.89. This fell to 0.63 when the E classifier had a higher priority (S$>$E$>$C). The score for the Stability criteria remained relatively unchanged since its priority wasn't altered. 
\begin{table}
\caption{Average Criteria Scores with Different Criteria Priorities.}
\label{table:changecriteria}
\begin{center}
\normalsize
\begin{tabular}{lccc}
\hline
                           & S        & E        & C        \\ \hline
Initial                         & 0.26 & 0.57 & 0.30 \\ \hline
S\textgreater{}C=E              & 0.70 & 0.63  & 0.83 \\
S\textgreater{}C\textgreater{}E & 0.69 & 0.56  & 0.89 \\
S\textgreater{}E\textgreater{}C & 0.72 & 0.76   & 0.63 \\
\hline
\end{tabular}
\end{center}
\vspace{-1em}
\end{table}

\para{Is combining gradient-based and gradient-free optimization beneficial?} \method-OPT uses a hybrid optimization scheme that blends derivative-free search with gradient-based local search. Fig. \ref{fig:gradvsnongrad} shows \method-OPT significantly outperforms using either solely gradient-based expected utility maximization (50 samples) or gradient-free optimization via ES (1000 samples). In the Shelf environment, \method-OPT achieves more than double the average success rate compared to the competing methods.

\begin{figure}
\begin{center}
\includegraphics[width=0.8\linewidth]{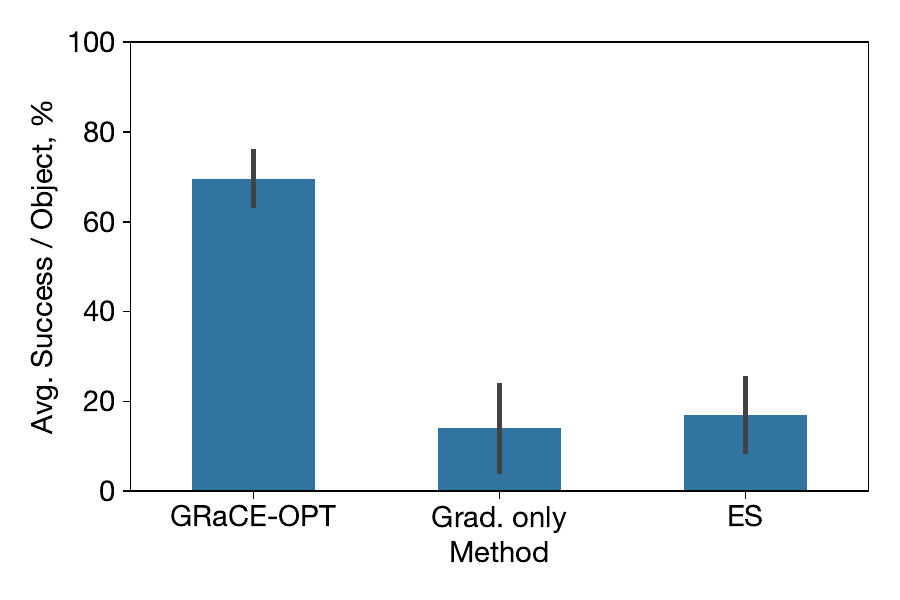}
\caption{\method-OPT significantly outperforms both  gradient ascent on the expected utility and the gradient-free Evolutionary Strategy (ES).}
\label{fig:gradvsnongrad}
\end{center}
\end{figure}


\para{What is the effect of changing the number of update steps $T$ and $K$ in \method-OPT?} In our experiments, we chose $T$ and $K$ based on our computational budget. Higher values of $T$ and $K$ lead to longer optimization times and improved grasp outcomes as shown in Fig. \ref{fig:ablatingTK}.

\begin{figure}
\centering
\begin{subfigure}{0.40\textwidth}
    \includegraphics[width=\linewidth]{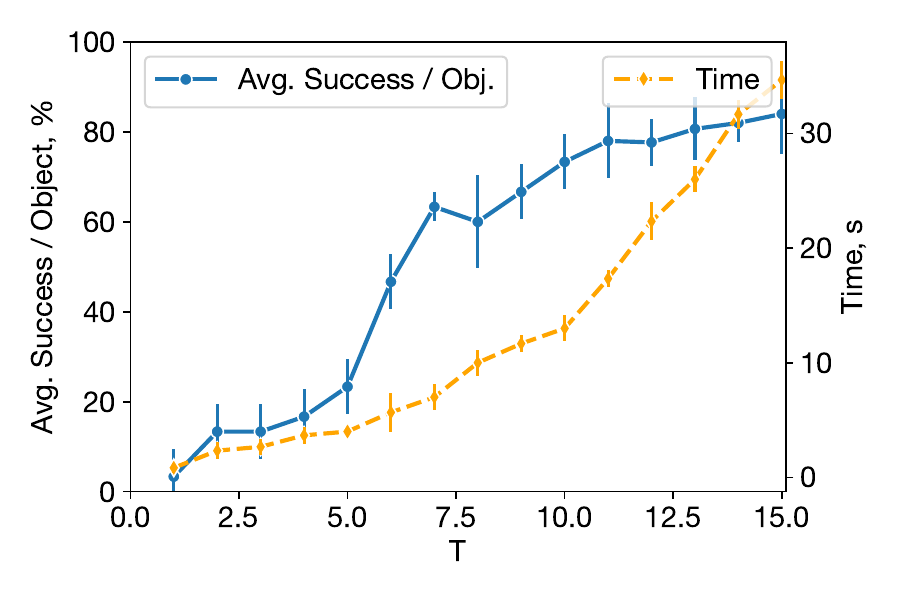}
    \caption{Varying $T$ while keeping $K=2$. }
    \label{fig:T_ablation}
\end{subfigure}
\hfill
\begin{subfigure}{0.40\textwidth}
    \includegraphics[width=\linewidth]{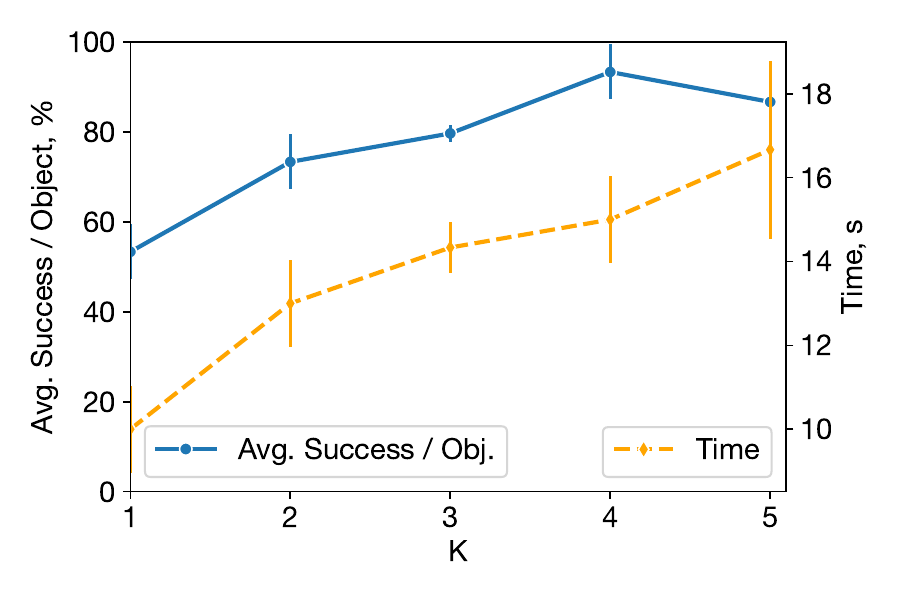}
    \caption{Varying $K$ while keeping $T=10$.}
    \label{fig:K_ablation}
\end{subfigure}
\caption{Increasing the number of outer update steps $T$ and inner gradient steps $K$ increases performance at the cost of longer computation times.}
\label{fig:ablatingTK}
\end{figure}
}
\section{Conclusions and Future Work}
\label{sec:discussion}


In this study, we introduced \method{}, a probabilistic hierarchical rank-based modular framework designed for optimizing robotic grasps based on multiple, often conflicting, criteria. We formulated these criteria as a  utility function based on the expected ranks of grasp; this takes into account the uncertainty inherent in many real-world grasping scenarios. In addition, we presented \method{}-OPT as a hybrid optimization technique to optimize grasps using both gradient-based and gradient-free methods. Our experimental evaluations show GRACE’s efficacy in generating high-quality grasps in complex, cluttered environments both in simulation and real-world experiments. 

\para{Limitations and Future Work.} The proposed work faces two primary limitations. Firstly, the reliance of \method{}-OPT on gradient calculations leads to increased computational costs with the addition of more criteria to the framework. A direct approach would be to improve the efficiency of the gradient computations, e.g., via approximations. Alternatively, one could explore exclusively gradient-free methods. Secondly, while \method{} has been tested with key criteria like reachability, singularity, and collision, the aspect of plannability remains unexplored. A lack of a feasible plan could render the final grasp unexecutable. A potential solution for future work involves integrating a planner-based classifier into the framework. Additionally, \method{} could be expanded to include more criteria, such as tactile feedback (e.g., \cite{taunyazov20event,taunyazov2021extended}) for grasping of soft or deformable objects. We believe \method's flexibility and modularity opens up promising avenues for future research.


\section*{Acknowledgements}

This research is supported by the National Research Foundation Singapore and DSO National Laboratories under the AI Singapore Programme (AISG Award No: AISG2-RP-2020-016).


\bibliographystyle{IEEEtran}
\balance
\bibliography{grace.bib}


\end{document}